\documentclass[letterpaper]{article} 
\usepackage[preprint]{aaai2027} 
\usepackage[hyphens]{url}  
\usepackage{graphicx} 
\usepackage{natbib}  
\usepackage{caption} 
\usepackage{algorithm}
\usepackage{algorithmic}

\usepackage{newfloat}
\usepackage{listings}
\DeclareCaptionStyle{ruled}{labelfont=normalfont,labelsep=colon,strut=off} 
\floatstyle{ruled}
\newfloat{listing}{tb}{lst}{}
\floatname{listing}{Listing}

\usepackage{booktabs}
\usepackage{multirow}
\usepackage{amsmath}
\usepackage{amssymb}
\newtheorem{definition}{Definition}
\newtheorem{proposition}{Proposition}
\usepackage{xspace}

\title{S-GAP: Stealthy Graph Attack Primitives for Graph Large Language Models}

\author{
    Wenda Li\textsuperscript{\rm 1},
    Tongya Zheng\textsuperscript{\rm 2},
    Shunyu Liu\textsuperscript{\rm 3},
    Hanyang Yuan\textsuperscript{\rm 1},
    Kaixuan Chen\textsuperscript{\rm 1}\\
    Yu Wang\textsuperscript{\rm 1},
    Yuchen Ying\textsuperscript{\rm 1},
    Haoze Jiang\textsuperscript{\rm 1},
    Huan Li\textsuperscript{\rm 1},
    Gang Chen\textsuperscript{\rm 1},
    Zujie Ren\textsuperscript{\rm 4,\textdagger}
}

\affiliations{
    \textsuperscript{\rm 1}Zhejiang University\quad
    \textsuperscript{\rm 2}Hangzhou City University\\
    \textsuperscript{\rm 3}Nanyang Technological University\quad
    \textsuperscript{\rm 4}Zhejiang Lab\\
    lwdup@zju.edu.cn, renzju@zju.edu.cn\\
    \textsuperscript{\textdagger}Corresponding author: Zujie Ren
}

\newcommand{\name}{S-GAP\xspace}

\begin{document}

\maketitle

\begin{abstract}

Graph Large Language Models~(GraphLLMs) enable powerful semantic understanding and reasoning over Text-Attributed Graphs~(TAGs), while their semantic reasoning may exhibit vulnerabilities to perturbations in node text and graph structure under adversarial attack evaluation.
Existing attacks typically construct textual and structural perturbations in separate stages guided by distinct realization intents. 
However, this separation results in \emph{attack-intent fragmentation}, causing semantic incoherence between the two perturbation types, thereby undermining attack stealth and making them easier for GraphLLMs to identify and disregard, ultimately weakening attack effectiveness.
To avoid \emph{attack-intent fragmentation}, we propose the \emph{Graph Attack Primitive}~(GAP), an atomic semantic representation that unifies textual and structural perturbations under a coherent attack intent. 
By carefully calibrating the semantic composition of a GAP, we can achieve strong attack effectiveness while maintaining sufficient attack stealth.
Furthermore, we introduce \name~(\textbf{S}tealthy \textbf{GAP}), a multi-agent framework that iteratively refines the semantic content of the GAP. 
The framework unifies GAP planning and realization and continuously refines this process through surrogate evaluation and textual feedback.
Extensive experiments across four datasets, three GraphLLMs, and two attack settings show that \name improves ASR over the strongest baselines by 7.02\% on average while producing more relationally plausible perturbations, exposing critical vulnerabilities in current GraphLLMs and motivating more robust defenses.




\end{abstract}

\noindent
\section{Introduction}


\begin{figure}[t]
    \centering
    \includegraphics[width=\columnwidth]{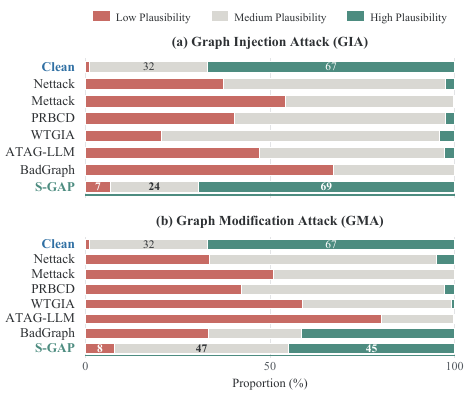}
    \caption{Relational-plausibility distributions on Cora. An LLM assesses each relation as having low, medium, or high plausibility, while Clean denotes original graph relations.}
    \label{fig:intro_edge_relevance}
\end{figure}

Text-Attributed Graphs (TAGs), which integrate graph structure with textual content, are widely used to represent citation networks~\cite{tang2024graphgpt}, product networks~\cite{fan2022modeling}, and other knowledge-rich networks~\cite{zhong2023comprehensive,ying2026graphscout}.
Recently, Graph Large Language Models~(GraphLLMs)~\cite{chen2024llaga,wang2024llms,zhang2026toward} have incorporated the powerful semantic understanding and reasoning capabilities of large language models into graph learning, enabling joint reasoning over node text and graph structure. Compared with conventional graph models, this integration allows GraphLLMs to contextualize node semantics using relational information, supporting richer and more context-sensitive graph reasoning.
However, this joint reliance also creates a coupled vulnerability: malicious text can distort node semantics, while manipulated relations can alter how those semantics are contextualized.
Understanding how this coupled vulnerability can be exploited by adversarial attacks is therefore critical to the safe and reliable deployment of GraphLLMs in knowledge-intensive and decision-sensitive applications~\cite{gao2025leveraging,dai2024comprehensive}.



Graph adversarial attacks systematically expose the vulnerabilities of GraphLLMs through budget-constrained perturbations to graph data and commonly operate in two settings: Graph Injection Attack~(GIA), which introduces adversarial nodes together with their incident edges, and Graph Modification Attack~(GMA), which alters existing graph elements.
Across these settings, conventional methods manipulate graph structure, vectorized node attributes, or both~\cite{zugner2018adversarial,zugner_adversarial_2019,ju2023let,ma2026graphtextack}. 
Attacks on TAGs impose an additional constraint: attribute perturbations must be realized as valid natural-language text while being coordinated with structural manipulations~\cite{lyu2025navigating,chen2026can,lei2024intruding}.
These methods typically construct textual and structural perturbations in separate stages, with each stage guided by a separately defined realization intent derived from the shared attack objective. 
For example, the structural stage~\cite{lyu2025navigating} is driven by relational criteria, whereas the textual stage~\cite{lei2024intruding} follows separately defined semantic constraints. Although both stages pursue the same ultimate attack objective, their stage-specific attack intents are defined by different realization criteria rather than explicitly unified. We refer to this separation as \emph{attack-intent fragmentation}.


Attack-intent fragmentation can lead to semantic incoherence between textual and structural perturbations, with the generated text failing to justify the manipulated relations and the structural perturbations failing to support the adversarial semantics introduced by the text. 
Such incoherent perturbations lack sufficient attack stealth, making them more likely to be judged semantically implausible or weakly relevant and subsequently disregarded by the LLM components of GraphLLMs, ultimately weakening attack effectiveness.
We further validate this observation by prompting an LLM to assess the plausibility of the relations produced by each attack and categorize them as low, medium, or high plausibility. As shown in Figure~\ref{fig:intro_edge_relevance}, existing attacks produce substantially larger proportions of low-plausibility relations than the clean graph under both attack settings, suggesting that their textual and structural perturbations frequently form semantically implausible relations. In contrast, our method \name yields plausibility distributions much closer to the clean reference, indicating that its perturbations better preserve the relational plausibility profile of the original graph and thus achieve stronger attack stealth.


To avoid attack-intent fragmentation, we propose the \emph{Graph Attack Primitive} (GAP), an atomic semantic representation that provides a coherent attack intent shared by textual and structural perturbations. By explicitly specifying the target semantics to preserve and the adversarial semantics to introduce, a GAP can avoid semantically implausible perturbations while retaining adversarial strength. However, balancing these competing semantic compositions is inherently target-dependent and difficult to achieve in a single generation step.
We therefore introduce \name~(\textbf{S}tealthy \textbf{GAP}), a multi-agent framework that iteratively refines GAPs through surrogate evaluation and textual feedback. The Planner Agent proposes a GAP for the target, which the Attack Agent instantiates as a concrete text--structure perturbation and evaluates using a surrogate GNN. 
The Main Agent controls the overall process and uses the target context and execution outcome to construct a textual feedback that guides the Planner Agent’s refinement in the next round.
Our contributions are summarized as follows:
\begin{itemize}
    \item 
    We identify \emph{attack-intent fragmentation} in adversarial attacks, where textual and structural perturbations guided by independent attack intents undermine stealth, and propose GAP, which provides a unified attack intent for both perturbation types and is expressively complete over the defined attack search space.
    \item 
    We introduce \name, a multi-agent framework for unified target-specific GAP planning and realization with iterative feedback-guided refinement through surrogate evaluation and textual feedback.
    \item We conduct comprehensive experiments on four datasets with three victim GraphLLMs under two attack settings. The results demonstrate superior attack effectiveness and stealth over existing methods, exposing fundamental vulnerabilities in the joint textual and structural reasoning of current GraphLLMs.
\end{itemize}

\begin{figure*}[t]
    \centering
    \includegraphics[width=\textwidth]{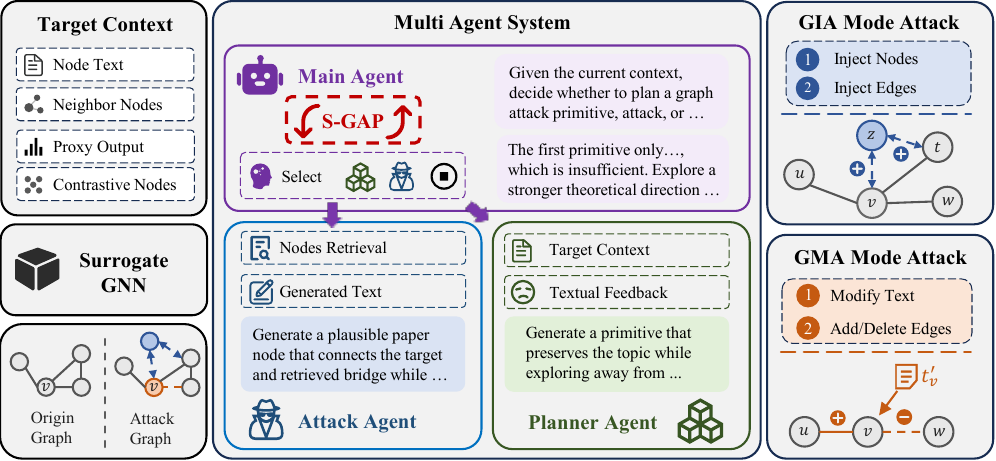}
    \caption{Overview of the proposed \name framework.}
    \label{fig:overview}
\end{figure*}

\section{Related Work}

\paragraph{Graph Adversarial Attacks.}
Graph adversarial attacks traditionally manipulate graph topology or vectorized node attributes. In the graph injection setting, attackers~\cite{chen2022understanding} introduce adversarial nodes together with their incident edges.
TDGIA~\citep{zou2021tdgia} jointly optimizes injected features and connectivity to increase their influence on target predictions, while reinforcement-learning-based approaches~\citep{sun2020adversarial,ju2023let,zhao2025highly} formulate node injection as a sequential decision process under different levels of model access.
In the graph modification setting, attacks~\cite{li2021adversarial,xu2022blindfolded} alter existing graph elements: Nettack~\citep{zugner2018adversarial} constructs targeted local perturbations by modifying edges and node attributes, Metattack~\citep{zugner_adversarial_2019} formulates graph poisoning as a meta-learning problem, and PRBCD~\citep{geisler2021robustness} improves the scalability of structural attacks through randomized block coordinate descent.
Despite their different threat models and optimization strategies, these methods operate primarily on graph topology and discrete or continuous feature representations, rather than directly constructing or editing the underlying natural-language text.

\paragraph{Attacks on Text-Attributed Graphs.}
Recent studies extend graph attacks from vectorized attributes to textual and semantic perturbations on TAGs.
WTGIA~\citep{lei2024intruding} pioneers text-level graph injection by converting adversarial feature patterns into interpretable textual nodes.
ATAG-LLM~\citep{lyu2025navigating} further employs LLMs to generate candidate injection texts and uses graph explanations to determine their injection positions in a black-box setting.
Beyond raw-text generation alone, GraphTextack~\citep{ma2026graphtextack} jointly searches the connectivity and semantic feature assignments of injected nodes through evolutionary optimization.
For graph modification, IMDGA~\citep{fan2026unveiling} combines graph-aware textual substitutions with edge pruning to improve attack effectiveness and stealthiness, while BadGraph~\citep{chen2026can} leverages graph priors and LLM reasoning to jointly perturb textual semantics and topology.
These studies establish textual semantics and graph structure as coupled attack surfaces.
However, their semantic coordination is embedded in particular generation procedures and perturbation mechanisms, such as node injection, textual substitution, or topology modification.
In contrast, \name introduces the Graph Attack Primitive as an explicit semantic interface between attack intent and concrete text--structure realization, enabling the same representation to guide both GIA and GMA.

\section{Background}

\paragraph{Text-Attributed Graphs.} A Text-Attributed Graph (TAG) is denoted by $\mathcal{G}=(\mathcal{V},\mathcal{E},\mathcal{T})$, where $\mathcal{V}$ is the node set, $\mathcal{E}\subseteq\mathcal{V}\times \mathcal{V}$ is the edge set, and $\mathcal{T}=\{t_i\mid v_i\in\mathcal{V}\}$ is the collection of node texts, with $t_i$ denoting the text associated with node $v_i$.
The graph structure is represented by an adjacency matrix $\mathbf{A}\in\{0,1\}^{|\mathcal{V}|\times|\mathcal{V}|}$, where $A_{ij}=1$ if $(v_i,v_j)\in\mathcal{E}$ and $A_{ij}=0$ otherwise.
In node classification, each node $v_i$ has a label $y_i\in\mathcal{Y}$, where $\mathcal{Y}$ denotes the label space \citep{lei2024intruding,chen2026can}.

\paragraph{Threat Model.}
We consider untargeted evasion attacks under a strict black-box setting. 
The attacker observes only the clean graph and training labels, but has no internal knowledge of or query access to the fixed victim model.
The goal is to change the target node's prediction to any other class. We consider two common graph perturbation settings~\citep{zou2021tdgia,ma2026graphtextack,chen2026can}.

\textbf{(1) Graph Injection Attack (GIA).}
The attacker injects a set of nodes $\mathcal{V}_v^{\mathrm{inj}}$ for a target node $v$, together with their associated texts $\mathcal{T}_v^{\mathrm{inj}}$ and incident edges $\mathcal{E}_v^{\mathrm{inj}}$.
The perturbed TAG is
\begin{equation}
\label{eq:gia-transformation}
\mathcal{G}'=\bigl(
\mathcal{V}\cup\mathcal{V}_v^{\mathrm{inj}},
\mathcal{E}\cup\mathcal{E}_v^{\mathrm{inj}},
\mathcal{T}\cup\mathcal{T}_v^{\mathrm{inj}}
\bigr).
\end{equation}
Each edge in $\mathcal{E}_v^{\mathrm{inj}}$ has at least one injected node as an endpoint, while the existing nodes, edges, and texts remain unchanged~\citep{zou2021tdgia,lei2024intruding,ma2026graphtextack}.

\textbf{(2) Graph Modification Attack (GMA).}
The attacker modifies the text of an existing node $v$ and its local connectivity, yielding a modified text $t'_v$, a set of added edges $\mathcal{E}_v^+$, and a set of removed edges $\mathcal{E}_v^-$.
The perturbed TAG is
\begin{equation}
\label{eq:gma-transformation}
\mathcal{G}'=\bigl(
\mathcal{V},
(\mathcal{E}\setminus\mathcal{E}_v^-)\cup\mathcal{E}_v^+,
(\mathcal T\setminus\{t_v\})\cup\{t'_v\}
\bigr).
\end{equation}
Each edge in $\mathcal{E}_v^+\cup\mathcal{E}_v^-$ has the target node $v$ as an endpoint, while the node set, all other node texts, and non-\(v\)-incident edges remain unchanged~\citep{zhang2025trustglm,chen2026can}.



\section{Methodology}

\subsection{Overview}

As illustrated in Figure~\ref{fig:overview}, \name is a multi-agent framework that iteratively refines GAPs using surrogate evaluation and textual feedback. 
The Main Agent coordinates the process by maintaining the attack state and deciding whether refinement should continue or terminate. 
At each iteration, the Planner Agent composes or refines a GAP. The Attack Agent then generates setting-specific text--structure perturbations from the GAP and evaluates them using a surrogate GNN.
The Main Agent then updates the attack state with the execution outcome and produces textual feedback to guide the Planner Agent's next refinement.

\subsection{Graph Attack Primitive}

\paragraph{Target context.}
Let $f_{\mathrm{sur}}$ be a surrogate GNN trained on the clean training split.
To reduce the combinatorial cost of endpoint selection over all $|\mathcal V|$ graph nodes, S-GAP restricts free attack-edge endpoints to a contrastive candidate pool $\mathcal D_v$ comprising the $M$ nodes in $\mathcal V$ with the lowest cosine similarity to $v$ under normalized text embeddings.
We construct the target context:
\begin{equation}
    \mathcal C_v=\left(t_v,\mathbf n_v^{\mathrm{text}},
    \mathbf p_v^{\mathrm{sur}},\mathbf d_v\right),
\end{equation}
where $t_v$ is the target text, $\mathbf n_v^{\mathrm{text}}$ is a bounded set of randomly sampled one-hop neighbors of $v$, $\mathbf p_v^{\mathrm{sur}}=f_{\mathrm{sur}}(\mathcal G,v)$ is the clean surrogate prediction, and $\mathbf d_v$ denotes a few nodes drawn from $\mathcal D_v$ and provided to the LLM as contrastive context, whose number is much smaller than $|\mathcal D_v|$.

\begin{definition}[Graph Attack Primitive]
\label{def:graph_attack_primitive}
A Graph Attack Primitive (GAP) is an atomic semantic representation of a coherent attack intent that serves as a general interface between the intent and its concrete text--structure realization.
Let $\mathcal P$ denote the primitive space, with $P_v\in\mathcal P$ representing a GAP associated with target node $v$. For an attack setting $m\in\{\mathrm{GIA},\mathrm{GMA}\}$, the realization map
\begin{equation}
    \mathcal I_m(v,P_v)
\end{equation}
returns the set of valid text--structure perturbations that realize the attack intent represented by $P_v$ under setting $m$.
\end{definition}
Beyond providing a unified attack intent for text–structure perturbations, another key advantage of GAP is its compatibility with both GMA and GIA. Existing methods are typically tailored to the perturbation mechanism of a single setting and therefore support either graph modification or graph injection. By separating attack intent from setting-specific realization, GAP accommodates both settings within a unified framework through their respective realization functions.

\begin{proposition}[Candidate Search-Space Completeness]
\label{prop:primitive_completeness}
For any candidate endpoint search space $\mathcal D_v\subseteq\mathcal V$ and graph budget $B$, let $\mathcal F_{v,m}^{\mathrm{cand}}(\mathcal D_v,B)$ denote all valid perturbations under setting $m$ whose free structural endpoints lie in $\mathcal D_v$ and whose graph cost is at most $B$. 
Assume that the LLM is text-complete and that retrieval is tie-free. Then
\begin{equation}
    \mathcal F_{v,m}^{\mathrm{cand}}(\mathcal D_v,B)
    \subseteq
    \bigcup_{P_v\in\mathcal P}\mathcal I_m(v,P_v).
    \label{eq:candidate_completeness}
\end{equation}
\end{proposition}
The proof is provided in Appendix. Proposition~\ref{prop:primitive_completeness} establishes candidate-space completeness under the text-completeness assumption. Accordingly, exhaustive optimization over all primitive realizations preserves the optimum within the candidate-restricted attack space. 

\paragraph{Stealth-oriented primitive representation.}
To improve attack stealthiness, \name represents each primitive using two semantic components:
\begin{equation}
    P_v=(\alpha_v,\beta_v),
\end{equation}
where $\alpha_v$ is the \emph{target anchor} that identifies the target semantics to preserve, and $\beta_v$ denotes the \emph{adversarial semantics} to introduce. 
The target anchor keeps the generated entity aligned with the target topic, while the adversarial semantics provide contrastive evidence against the current prediction.

The GAP provides a compact and reusable semantic specification that establishes a unified attack direction for text--structure perturbations, while its concrete realization as nodes and edges is determined by the system during instantiation.
Since $P_v$ remains a target-specific decision variable, identifying an effective $P_v$ motivates the iterative multi-agent optimization framework presented next.




\subsection{Framework of Stealthy Graph Attack Primitive}

For target $v$ under attack setting $m$, the Planner, Attack, and Main Agents jointly optimize $P_v$ through a recurrent cycle of primitive planning, attack realization followed by evaluation, and feedback synthesis.

\paragraph{Primitive planning.}
At round $r$, the Planner Agent generates a primitive from the target context and textual feedback produced by the Main Agent:
\begin{equation}
    P_v^{(r)}=\mathrm{PlannerAgent}(\mathcal C_v,g_{v,m}^{(r-1)}).
\end{equation}
Here, $g_{v,m}^{(r-1)}$ denotes the textual feedback generated by the Main Agent described below, and $g_{v,m}^{(-1)}=\varnothing$.

\paragraph{Attack realization and evaluation.}
Given $P_v^{(r)}$, the Attack Agent realizes it as $K$ setting-specific executable text--structure perturbations $\{\Delta_{v,i,m}^{(r)}\}_{i=1}^{K}$.
To compare their attack efficacy, the Attack Agent evaluates each perturbation on the surrogate GNN using a confidence-drop score, defined as the reduction in confidence assigned to the class predicted on the clean graph:
\begin{equation}
    J_{v,i,m}^{(r)}=
    f_{\mathrm{sur}}(\mathcal G,v)_{c_v}-
    f_{\mathrm{sur}}\left(\mathcal G\oplus_m\Delta_{v,i,m}^{(r)},v\right)_{c_v},
\end{equation}
where $c_v=\arg\max_c f_{\mathrm{sur}}(\mathcal{G},v)_c$, and $\oplus_m$ denotes the setting-specific graph transformation defined in Eqs.~\eqref{eq:gia-transformation} and~\eqref{eq:gma-transformation}.
A larger $J_{v,i,m}^{(r)}$ indicates a greater reduction in the surrogate confidence of the clean prediction.

The Attack Agent returns the resulting perturbations and their confidence-drop scores as its round output:
\begin{equation}
\begin{aligned}
    O_{v,m}^{(r)}
    &=\mathrm{AttackAgent}_m(\mathcal C_v,P_v^{(r)})\\
    &=\left\{\left(\Delta_{v,i,m}^{(r)},J_{v,i,m}^{(r)}\right)\right\}_{i=1}^{K}.
\end{aligned}
\end{equation}

\paragraph{Textual feedback and iterative refinement.}
The Main Agent controls the refining optimization process by maintaining a context history $\mathcal H_{v,m}^{(r)}$ and selecting each subsequent operation.
This history records the planning and execution results from previous rounds, prior operation instructions, the current best perturbation and edit plan, and the remaining budget.
Let $\mathrm{Out}_{v,m}^{(r)}$ denote the output returned by the agent invoked in round $r$:
\begin{equation}
    \mathrm{Out}_{v,m}^{(r)}=
    \begin{cases}
        P_v^{(r)}, & \text{if the Planner Agent is invoked},\\
        O_{v,m}^{(r)}, & \text{if the Attack Agent is invoked}.
    \end{cases}
\end{equation}

Given the current output and the preceding context history, the Main Agent selects the next operation:
\begin{equation}
\begin{aligned}
    \eta_{v,m}^{(r)}&=\mathrm{MainAgent}\left(\mathrm{Out}_{v,m}^{(r)},\mathcal H_{v,m}^{(r-1)}\right),\\
    \eta_{v,m}^{(r)}&\in\left\{\mathrm{Plan},\mathrm{Attack},\mathrm{Stop}\right\}.
\end{aligned}
\end{equation}
The current output and selected operation are then incorporated into $\mathcal H_{v,m}^{(r)}$.
When $\eta_{v,m}^{(r)}=\mathrm{Plan}$, the Planner Agent receives a planner-facing GAP context $\widetilde{\mathcal C}_{v,m}^{(r)}$ and the Main Agent's textual assessment $e_{v,m}^{(r)}$ of the current result as textual feedback:
\begin{equation}
    g_{v,m}^{(r)}=
    \left(\widetilde{\mathcal C}_{v,m}^{(r)},e_{v,m}^{(r)}\right).
\end{equation}
The textual feedback $g_{v,m}^{(r)}$ provides a natural-language update direction for the primitive components, guiding the revision from $P_v^{(r)}$ to $P_v^{(r+1)}$.
When $\eta_{v,m}^{(r)}=\mathrm{Attack}$, the Attack Agent executes an existing primitive and returns its outcome.
When $\eta_{v,m}^{(r)}=\mathrm{Stop}$, optimization for the current target terminates.
This Main-Agent-controlled adaptive decision process allocates fewer primitive rounds to easier targets and more rounds to harder ones, subject to the maximum round budget.

\subsection{GAP-guided Perturbation Construction}
GIA and GMA share the same GAP optimization process but use setting-specific perturbation construction.
For each $i\in\{1,\ldots,K\}$, the Attack Agent constructs the $i$-th executable text--structure perturbation from a fixed primitive $P_v$ by retrieving a setting-specific endpoint plan and generating its corresponding textual realization:
\begin{align}
    \Omega_{v,i,m}&=\operatorname{Retrieve}_m\left(\mathcal C_v,\mathcal D_v,P_v\right),\\
    T_{v,i,m}&=\operatorname{Generate}_m\left(\mathcal C_v,P_v,\Omega_{v,i,m}\right).
\end{align}
Here, $\Omega_{v,i,m}$ specifies the retrieved endpoints, while $T_{v,i,m}$ denotes the generated textual realization.
Guided by the same GAP, the two perturbations convey consistent semantics and maintain relational plausibility, resulting in improved attack stealthiness.
Their setting-specific composition into $\Delta_{v,i,m}$ is defined below.

\noindent
\textbf{Graph Injection Attack Realization.}
Under GIA, $\Omega_{v,i,\mathrm{GIA}}$ specifies the existing endpoints retrieved from $\mathcal D_v$, while $T_{v,i,\mathrm{GIA}}=\mathcal T_{v,i}^{\mathrm{inj}}$ contains the texts generated for the injected nodes.
Together, they define the perturbation
\begin{equation}
    \Delta_{v,i,\mathrm{GIA}}=
    \left(\mathcal V_{v,i}^{\mathrm{inj}},
    \mathcal T_{v,i}^{\mathrm{inj}},
    \mathcal E_{v,i}^{\mathrm{inj}}\right),
\end{equation}
where $\mathcal V_{v,i}^{\mathrm{inj}}$ contains the instantiated nodes, $\mathcal T_{v,i}^{\mathrm{inj}}$ denotes the corresponding generated node texts, and $\mathcal E_{v,i}^{\mathrm{inj}}$ connects these nodes to the endpoints specified by $\Omega_{v,i,\mathrm{GIA}}$.
The attack is applied as
$\mathcal{G}'=\mathcal{G}\oplus_{\mathrm{GIA}}\Delta_{v,i,\mathrm{GIA}}$,
following the graph transformation in Eq.~\eqref{eq:gia-transformation}.

\noindent
\textbf{Graph Modification Attack Realization.}
Under GMA, $\Omega_{v,i,\mathrm{GMA}}$ specifies addition endpoints selected from $\mathcal D_v$ and removal endpoints selected from the existing neighbors of $v$, while $T_{v,i,\mathrm{GMA}}$ is the generated rewrite of the target text.
Together, they define the perturbation
\begin{equation}
\Delta_{v,i,\mathrm{GMA}}
=
\left(
T_{v,i,\mathrm{GMA}},
\mathcal E_{v,i}^{+},
\mathcal E_{v,i}^{-}
\right),
\end{equation}
where $\mathcal E_{v,i}^{+}$ connects $v$ to the addition endpoints specified by $\Omega_{v,i,\mathrm{GMA}}$, and $\mathcal E_{v,i}^{-}$ removes its edges to the specified existing neighbors.
The attack is applied as
$\mathcal{G}'=\mathcal{G}\oplus_{\mathrm{GMA}}\Delta_{v,i,\mathrm{GMA}}$,
following the graph transformation in Eq.~\eqref{eq:gma-transformation}.

\section{Experiments}

\subsection{Experimental Setup}

\paragraph{Datasets.}
We conduct experiments on four text-attributed graphs: Cora~\cite{mccallum2000automating}, PubMed~\cite{sen2008collective}, OGBN-Arxiv~\cite{hu2020open}, and a subset of OGBN-Products~\cite{hu2020open}.
More details and statistics of the datasets are provided in Appendix.

\begin{table*}[t]
\centering
\footnotesize
\setlength{\tabcolsep}{1mm}
\begin{tabular}{lll*{9}{c}}
\toprule
Setting & Dataset & Metric/Model & Clean & Nettack & Mettack & PRBCD & G2A2C & WTGIA & ATAG-LLM & BadGraph & \name \\
\midrule
\multirow{16}{*}{GIA}
& \multirow{4}{*}{Cora} & LLaGA & -- & 3.01 & 3.01 & 3.01 & 1.72 & 2.37 & 1.94 & 3.23 & \textbf{6.02} \\
& & RGLM & -- & 1.26 & 1.68 & 0.84 & 1.26 & 0.63 & 0.84 & 1.05 & \textbf{3.15} \\
& & TEA-GLM & -- & 2.93 & 3.66 & 1.95 & 4.15 & 4.39 & 2.20 & 5.12 & \textbf{7.07} \\
& & High Plaus. & 66.88 & 2.40 & 0.20 & 2.41 & -- & 4.25 & 2.86 & 0.18 & \textbf{69.43} \\
\cmidrule(lr){2-12}
& \multirow{4}{*}{PubMed} & LLaGA & -- & 1.78 & 1.89 & 2.09 & 1.83 & 2.00 & 1.94 & 2.00 & \textbf{2.79} \\
& & RGLM & -- & 0.47 & 0.44 & 0.69 & 0.64 & 0.75 & 0.66 & 0.64 & \textbf{1.47} \\
& & TEA-GLM & -- & 0.19 & 0.08 & 0.25 & 0.33 & 0.22 & 0.44 & 0.14 & \textbf{0.58} \\
& & High Plaus. & 87.12 & 2.10 & 2.92 & 4.03 & -- & 6.80 & 2.40 & 1.30 & \textbf{61.19} \\
\cmidrule(lr){2-12}
& \multirow{4}{*}{OGBN-Arxiv} & LLaGA & -- & 4.48 & 4.40 & 4.89 & 4.26 & 4.95 & 5.11 & 4.20 & \textbf{6.92} \\
& & RGLM & -- & 1.26 & 1.88 & 1.64 & 1.75 & 1.67 & 1.78 & 1.34 & \textbf{3.25} \\
& & TEA-GLM & -- & 9.88 & 9.58 & 10.99 & 11.37 & 10.30 & 10.99 & 8.42 & \textbf{12.69} \\
& & High Plaus. & 63.90 & 0.00 & 3.20 & 0.25 & -- & 0.70 & 1.30 & 0.00 & \textbf{67.57} \\
\cmidrule(lr){2-12}
& \multirow{4}{*}{OGBN-Products} & LLaGA & -- & 2.17 & 2.09 & 2.14 & 2.62 & 3.10 & 3.72 & 2.19 & \textbf{4.07} \\
& & RGLM & -- & 0.29 & 0.18 & 0.29 & 0.34 & 0.84 & 1.05 & 0.18 & \textbf{2.04} \\
& & TEA-GLM & -- & 3.59 & 2.94 & 3.74 & 4.09 & 3.78 & 3.19 & 3.47 & \textbf{5.53} \\
& & High Plaus. & 52.43 & 0.00 & 2.24 & 0.15 & -- & 0.25 & 0.30 & 1.88 & \textbf{26.78} \\
\midrule
\multirow{16}{*}{GMA}
& \multirow{4}{*}{Cora} & LLaGA & -- & 38.28 & 23.23 & 26.67 & 3.01 & 2.37 & 17.63 & 19.14 & \textbf{44.95} \\
& & RGLM & -- & 28.36 & 10.71 & 22.48 & 1.89 & 0.63 & 9.45 & 21.22 & \textbf{49.16} \\
& & TEA-GLM & -- & 10.24 & 9.76 & 9.51 & 1.71 & 4.39 & 6.59 & 21.95 & \textbf{50.73} \\
& & High Plaus. & 66.88 & 4.89 & 0.09 & 2.77 & -- & 0.81 & 0.30 & 41.61 & \textbf{44.92} \\
\cmidrule(lr){2-12}
& \multirow{4}{*}{PubMed} & LLaGA & -- & 27.34 & 1.80 & 13.28 & 3.33 & 4.23 & 7.98 & 26.30 & \textbf{49.01} \\
& & RGLM & -- & 8.06 & 0.42 & 5.01 & 0.94 & 1.75 & 2.02 & 29.42 & \textbf{47.76} \\
& & TEA-GLM & -- & 0.92 & 0.08 & 0.92 & 0.14 & 0.44 & 0.56 & 42.50 & \textbf{56.89} \\
& & High Plaus. & 87.12 & 6.35 & 2.67 & 4.44 & -- & 12.66 & 9.70 & 46.53 & \textbf{46.90} \\
\cmidrule(lr){2-12}
& \multirow{4}{*}{OGBN-Arxiv} & LLaGA & -- & 9.73 & 4.62 & 13.32 & 5.30 & 9.53 & 6.68 & 20.25 & \textbf{25.19} \\
& & RGLM & -- & 3.23 & 1.72 & 5.51 & 1.75 & 2.96 & 2.82 & 21.54 & \textbf{23.45} \\
& & TEA-GLM & -- & 11.34 & 9.49 & 13.49 & 9.94 & 12.93 & 11.07 & 15.63 & \textbf{30.33} \\
& & High Plaus. & 63.90 & 29.00 & 1.71 & 0.50 & -- & 0.30 & 0.03 & 24.47 & \textbf{42.05} \\
\cmidrule(lr){2-12}
& \multirow{4}{*}{OGBN-Products} & LLaGA & -- & 8.78 & 2.25 & 11.67 & 7.57 & 7.92 & 10.89 & 34.17 & \textbf{49.08} \\
& & RGLM & -- & 3.98 & 0.16 & 6.10 & 1.57 & 3.59 & 5.40 & 28.50 & \textbf{32.17} \\
& & TEA-GLM & -- & 4.05 & 3.19 & 3.50 & 3.87 & 3.74 & 3.93 & 27.48 & \textbf{30.11} \\
& & High Plaus. & 52.43 & 4.48 & 1.48 & 0.15 & -- & 0.42 & 0.00 & 22.56 & \textbf{35.30} \\
\bottomrule
\end{tabular}
\caption{Overall attack effectiveness and relational plausibility under GIA and GMA. 
We report ASR (\%) for attack effectiveness and High-Plaus. (\%) for relational plausibility.
Clean denotes original graph relations; bold indicates the best attack result.}
\label{tab:overall_results}
\end{table*}

\begin{table}[t]
\centering
\footnotesize
\renewcommand{\arraystretch}{0.9}
\setlength{\tabcolsep}{3.2pt}
\begin{tabular}{lllrrr}
\toprule
Setting & Dataset & Variant & LLaGA & RGLM & TEA-GLM \\
\midrule
\multirow{8}{*}{GIA}
& \multirow{4}{*}{Cora} & w/o score & 2.58 & 1.05 & 3.66 \\
& & w/o GAP & 3.44 & 1.47 & 4.15 \\
& & w/o feedback & 3.66 & 1.26 & 6.83 \\
& & \textbf{\name} & \textbf{6.02} & \textbf{3.15} & \textbf{7.07} \\
\cmidrule(lr){2-6}
& \multirow{4}{*}{PubMed} & w/o score & 2.09 & 0.80 & 0.22 \\
& & w/o GAP & 2.17 & 1.02 & 0.28 \\
& & w/o feedback & 2.68 & 1.05 & 0.33 \\
& & \textbf{\name} & \textbf{2.79} & \textbf{1.47} & \textbf{0.58} \\
\midrule
\multirow{8}{*}{GMA}
& \multirow{4}{*}{Cora} & w/o score & 30.32 & 25.00 & 33.66 \\
& & w/o GAP & 23.23 & 22.06 & 26.59 \\
& & w/o feedback & 27.31 & 28.15 & 31.71 \\
& & \textbf{\name} & \textbf{44.95} & \textbf{49.16} & \textbf{50.73} \\
\cmidrule(lr){2-6}
& \multirow{4}{*}{PubMed} & w/o score & 25.96 & 17.92 & 25.71 \\
& & w/o GAP & 22.44 & 19.89 & 27.65 \\
& & w/o feedback & 22.52 & 18.53 & 25.96 \\
& & \textbf{\name} & \textbf{49.01} & \textbf{47.76} & \textbf{56.89} \\
\bottomrule
\end{tabular}
\caption{Ablation of surrogate scoring, GAP, and textual feedback. Results are reported as ASR (\%).}
\label{tab:feedback_ablation}
\end{table}

\paragraph{Victim Models and Baselines.}
We evaluate transfer attacks against three GraphLLMs: LLaGA~\cite{chen2024llaga}, RGLM~\cite{zhang2026toward}, and TEA-GLM~\cite{wang2024llms}.
We group the baselines according to their original attack settings.
The native GIA baselines are G2A2C~\cite{ju2023let}, WTGIA~\cite{lei2024intruding}, and ATAG-LLM~\cite{lyu2025navigating}, whereas the native GMA baselines are Nettack~\cite{zugner2018adversarial}, Mettack~\cite{zugner_adversarial_2019}, PRBCD~\cite{geisler2021robustness}, and BadGraph~\cite{chen2026can}.
See the appendix for model and baseline details.

\paragraph{Attack Settings.}
For GIA~\cite{zou2021tdgia,lei2024intruding,ma2026graphtextack}, the attacker injects one textual node for each target and adds at most two incident edges.
For GMA~\citep{zhang2025trustglm,chen2026can}, the attacker may make similarity-constrained revisions to the target text and add or remove at most two target-incident edges.

\paragraph{Evaluation metrics.}

We measure attack effectiveness by Attack Success Rate (ASR) over targets correctly classified on the clean graph~\citep{zhang2025trustglm,lyu2025navigating}, where higher ASR indicates stronger attacks. Attack stealthiness is measured by the relational plausibility of induced edges, scored from 1 to 6 by a post-hoc LLM using endpoint texts. Scores of 5--6 denote High plausibility, whose proportion increases with stealthiness; original graph relations serve as a reference. Full results for all plausibility levels are reported in the appendix.
\subsection{Overall Attack Evaluation}

\paragraph{Attack effectiveness.}
Table~\ref{tab:overall_results} shows that \name achieves the highest ASR in all 24 cases, outperforming the strongest baseline by 12.78 and 1.25 percentage points on average under GMA and GIA, respectively. GMA is generally more effective because it directly modifies target nodes, whereas GIA is more practical but influences predictions indirectly through injected nodes. Baselines perform better in their native settings, suggesting limited adaptability across attack settings. In contrast, the unified GAP representation enables \name to naturally adapt to both GIA and GMA, achieving consistently strong performance in both settings.
Nettack, Mettack, PRBCD, and G2A2C mainly optimize structures or numerical features using GNN-oriented objectives, without explicitly modeling raw-text semantics. WTGIA, ATAG-LLM, and BadGraph incorporate textual information but depend on predefined features, similarity proxies, or fixed attack strategies, limiting the coordination between text and structure. In contrast, \name directly models semantic attack intentions and uses execution feedback to jointly refine textual and structural perturbations, thereby avoiding reliance on numerical objectives or fixed proxies.

\begin{table}[t]
\centering
\small
\setlength{\tabcolsep}{4pt}
\begin{tabular}{llrrr}
\toprule
Dataset & Variant & LLaGA & RGLM & TEA-GLM \\
\midrule
\multirow{3}{*}{Cora} & Text & 15.48 & 16.81 & 41.71 \\
& Structure & 27.10 & 15.34 & 8.29 \\
& \textbf{\name} & \textbf{44.95} & \textbf{49.16} & \textbf{50.73} \\
\midrule
\multirow{3}{*}{PubMed} & Text & 16.54 & 21.44 & 38.62 \\
& Structure & 13.59 & 4.96 & 0.67 \\
& \textbf{\name} & \textbf{49.01} & \textbf{47.76} & \textbf{56.89} \\
\bottomrule
\end{tabular}
\caption{Comparison of \name with variants using only textual or structural perturbations under GMA.}
\label{tab:text_structure_ablation}
\end{table}

\begin{figure*}[t]
    \centering
    \includegraphics[width=\textwidth]{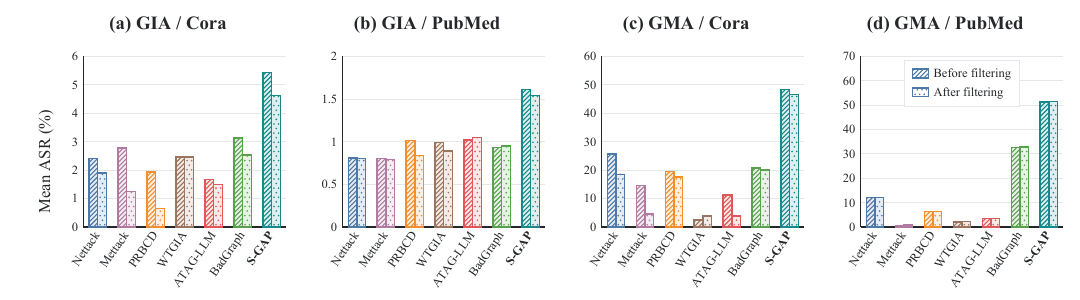}
    \caption{Victim-averaged ASR (\%) before and after reverting edge perturbations assigned relational-plausibility scores of 1--2.}
    \label{fig:semantic_filtering}
\end{figure*}


\paragraph{Attack stealthiness.}
Table~\ref{tab:overall_results} shows that \name achieves the highest relational plausibility across all datasets and settings. Structure-oriented attacks such as Nettack, Mettack, and PRBCD generally exhibit low plausibility because their edge perturbations lack semantic constraints; for example, Nettack achieves strong GMA ASR on Cora but only 4.89\% High-Plaus. 
Text-aware methods such as WTGIA, ATAG-LLM, and BadGraph achieve better plausibility by generating natural-language perturbations. However, their separately optimized text and structure or fixed influencer strategies limit semantic alignment. As low-plausibility relations may be ignored by GraphLLMs, we further examine their impact on attack effectiveness in the next subsection.

\subsection{Robustness to Relational Plausibility Filtering}
To determine whether attack success arises from plausible adversarial evidence or anomalous relational shortcuts, we apply a relational-plausibility filter that reverts edge perturbations assigned scores of 1--2 while retaining the textual perturbations. Several baselines degrade substantially on Cora, indicating reliance on semantically unjustified relations. In contrast, changes on PubMed are minor because low-plausibility relations are rare. \name retains at least 85.0\% of its ASR and remains the strongest attack after filtering, showing that its effectiveness primarily arises from plausible text--structure interactions. Its slight improvement under PubMed GMA further suggests that some implausible relations may even interfere with the attack. Complete plausibility-score distributions and victim-specific post-filtering results are provided in the appendix.

\begin{figure}[H]
    \centering
    \includegraphics[width=\columnwidth]{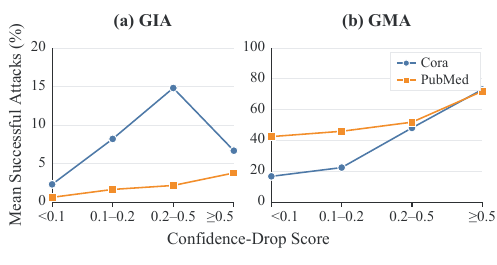}
    \caption{Victim-averaged proportions of successful attacks across confidence-drop score bins.}
    \label{fig:confidence_drop_score}
\end{figure}

\subsection{Quantitative Confidence-Drop Analysis}
To examine whether the confidence-drop score identifies perturbations likely to succeed on target GraphLLMs, we group attacks by their final scores and measure the victim-averaged proportion of successful attacks in each bin. 
Figure~\ref{fig:confidence_drop_score} shows that higher scores generally correspond to larger success proportions. The relatively low proportions reflect the prediction gap between the surrogate model and target GraphLLMs, while the overall trend confirms that the score remains informative of attack success.

\subsection{Ablation Studies}

\paragraph{Effects of confidence-drop scoring, GAP, and feedback.}
We compare \name with three variants: \emph{w/o score} removes confidence-drop scoring from the Attack Agent, \emph{w/o GAP} removes the GAP entirely and directly generates candidate attack entities from the target context, and \emph{w/o feedback}  removes textual feedback, with no refinement of GAPs. 
Table~\ref{tab:feedback_ablation} shows that removing any component consistently reduces ASR. Under GMA, the victim-averaged ASR decreases by 18.62--28.88 percentage points. Under GIA, \name also remains the best-performing configuration, although with smaller margins, demonstrating the benefit of the complete design under matched execution budgets.

\paragraph{Text--structure complementarity.}
Table~\ref{tab:text_structure_ablation} compares \name with variants using only textual or structural perturbations under GMA. 
\name achieves a victim-averaged ASR of 49.75\%, compared with 25.10\% and 11.66\% for the two variants, respectively.
These results demonstrate that the two perturbation modalities provide complementary attack signals and are more effective when used jointly.

\FloatBarrier
\section{Conclusion}
We investigate vulnerabilities in GraphLLMs’ joint reasoning over textual attributes and graph structure in TAGs. Existing attacks may underexpose these vulnerabilities because \emph{attack-intent fragmentation} misaligns textual and structural perturbations, weakening stealth and effectiveness. We therefore propose GAP to unify attack intent and develop \name to iteratively refine and realize GAPs through multi-agent planning, surrogate evaluation, and textual feedback. Across all 24 combinations of four datasets, three GraphLLMs, and two attack settings, \name performs best in every case, outperforming the strongest baselines by 7.02\% on average while producing more relationally plausible perturbations.

\noindent
\textbf{Social Impact}. This work reveals underexplored security risks in GraphLLMs arising from their joint reliance on text and graph structure. Our findings can support practical security evaluation and defense development for safer deployment in high-impact applications.

\bibliography{aaai2027}

\end{document}